\documentclass[sigconf, anonymous=false]{acmart}

\AtBeginDocument{%
  }

\setcopyright{acmlicensed}
\copyrightyear{2024}
\acmYear{2024}
\acmDOI{XXXXXXX.XXXXXXX}

\acmConference[Conference acronym 'XX]{Make sure to enter the correct
  conference title from your rights confirmation emai}{June 03--05,
  2018}{Woodstock, NY}

\acmISBN{978-1-4503-XXXX-X/18/06}

\usepackage{subcaption} % For subfigures
\usepackage{tikz} % For creating graphics programmatically
\usepackage{pgfplots}
\pgfplotsset{compat=1.16} 
\usepackage{algorithm}
\usepackage{algorithmic}
\usepackage{multirow}
\usepackage{hhline}
\usepackage{amsmath}
\usepackage{graphicx}
\usepackage{float}

\begin{document}
\sloppy

\title{Conformalized Selective Regression}
\author{Anna Sokol}
%\email{asokol@nd.edu}
\affiliation{%
  \institution{University of Notre Dame}
%  \city{Notre Dame}
%  \state{Indiana}
  \country{USA}
}

\author{Nuno Moniz}
%\email{nmoniz@nd.edu}
\affiliation{%
  \institution{University of Notre Dame}
%  \city{Notre Dame}
%  \state{Indiana}
  \country{USA}
}

\author{Nitesh Chawla}
%\email{nchawla@nd.edu}
\affiliation{%
  \institution{University of Notre Dame}
%  \city{Notre Dame}
%  \state{Indiana}
  \country{USA}
}

\begin{abstract}
Should prediction models always deliver a prediction? In the pursuit of maximum predictive performance, critical considerations of reliability %and fairness 
are often overshadowed, particularly when it comes to the role of uncertainty. Selective regression, also known as the ``reject option,'' allows models to abstain from predictions in cases of considerable uncertainty. Initially proposed seven decades ago, approaches to selective regression have mostly focused on distribution-based proxies for measuring uncertainty, particularly conditional variance. However, this focus neglects the significant influence of model-specific biases on performance. In this paper, we propose a novel approach to selective regression by leveraging conformal prediction, which provides grounded confidence measures for individual predictions based on model-specific biases. In addition, we propose a standardized evaluation framework to allow proper comparison of selective regression approaches. Via an extensive experimental approach, we demonstrate how our proposed approach, conformalized selective regression, presents an advantage over multiple state-of-the-art comparison models.

\end{abstract}

\begin{CCSXML}
<ccs2012>
 <concept>
  <concept_id>10010147.10010257.10010282.10010292</concept_id>
  <concept_desc>Computing methodologies~Machine learning</concept_desc>
  <concept_significance>500</concept_significance>
 </concept>
 <concept>
  <concept_id>10010147.10010257.10010293.10010307</concept_id>
  <concept_desc>Computing methodologies~Machine learning approaches</concept_desc>
  <concept_significance>300</concept_significance>
 </concept>
 <concept>
  <concept_id>10010147.10010257.10010321</concept_id>
  <concept_desc>Computing methodologies~Learning paradigms</concept_desc>
  <concept_significance>100</concept_significance>
 </concept>
 <concept>
  <concept_id>10003752.10010070.10010071.10010073</concept_id>
  <concept_desc>Theory of computation~Machine learning theory</concept_desc>
  <concept_significance>100</concept_significance>
 </concept>
</ccs2012>
\end{CCSXML}

\ccsdesc[500]{Computing methodologies~Machine learning}
\ccsdesc[300]{Computing methodologies~Machine learning approaches}
\ccsdesc[100]{Computing methodologies~Learning paradigms}
\ccsdesc[100]{Theory of computation~Machine learning theory}

\keywords{selective regression, conformal prediction, machine learning, uncertainty}

\received{20 February 2007}
\received[revised]{12 March 2009}
\received[accepted]{5 June 2009}

\maketitle

\section{Introduction}

\begin{figure}[ht!]
    \centering
    \begin{tikzpicture}
        \begin{axis}[
            xlabel={Coverage},
            ylabel={Error},
            xmin=0, xmax=1,
            ymin=0, ymax=1,
            xtick={0, 0.2, 0.4, 0.6, 0.8, 1},
            ytick={0, 0.2, 0.4, 0.6, 0.8, 1},
            grid style=none,
            width=0.45\linewidth,  % Reduced width
            height=0.45\linewidth, % Reduced height
            title={},
            legend columns=-1,
            legend entries={Model 1, Model 2},
            legend style={draw=none, font=\footnotesize}, % Smaller legend font
            legend to name=globalLegend, 
        ]
        \addplot[thick, color=orange, fill=orange, fill opacity=0.3] coordinates {
            (0.00, 0.25) (0.11, 0.26) (0.22, 0.27) (0.33, 0.29) (0.44, 0.30) (0.55, 0.31) (0.67, 0.32) (0.78, 0.33) (0.89, 0.35) (1.00, 0.45)
        }\closedcycle;
        \addplot[thick, color=blue, fill=blue, fill opacity=0.3] coordinates {
            (0.0, 0.04) (0.11, 0.04) (0.22, 0.06) (0.33, 0.07) (0.44, 0.08) (0.55, 0.11) (0.6, 0.14) (0.67, 0.20) (0.7, 0.22) (0.75, 0.26) (0.8, 0.30) (0.85, 0.35) (0.9, 0.52) (1.0, 0.80)
        }\closedcycle;
        \end{axis}
    \end{tikzpicture}
    \quad
    \begin{tikzpicture}
        \begin{axis}[
            xlabel={Coverage},
            ylabel={},
            xmin=0, xmax=1,
            ymin=0, ymax=1,
            xtick={0, .2, .4, .6, .8, 1},
            ytick=\empty,
            grid=none,
            width=0.45\linewidth, % Reduced width
            height=0.45\linewidth, % Reduced height
            title={},
            legend columns=-1,
            legend entries={Model 1, Model 2},
            legend style={draw=none, font=\footnotesize}, % Smaller legend font
            legend to name=globalLegend, 
        ]
        \addplot[color=orange, thick] coordinates {
            (0.00, 0.25) (0.11, 0.26) (0.22, 0.27) (0.33, 0.29) (0.44, 0.30) (0.55, 0.31) (0.67, 0.32) (0.78, 0.33) (0.89, 0.35) (1.00, 0.45)
        };
        \addplot[color=blue, thick] coordinates {
            (0.0, 0.04) (0.11, 0.04) (0.22, 0.06) (0.33, 0.07) (0.44, 0.08) (0.55, 0.11) (0.6, 0.14) (0.67, 0.20) (0.7, 0.22) (0.75, 0.26) (0.8, 0.30) (0.85, 0.35) (0.9, 0.52) (1.0, 0.80)
        };
        \node[circle, draw=darkgray, fill=darkgray, inner sep=1.4pt] at (axis cs:0.87, 0.34) {}; % Smaller point
        \draw[dashed, darkgray] (axis cs:1,0) -- (axis cs:0.87,0.35);
        \draw[dashed, darkgray] (axis cs:1,0) circle[radius=0.8cm];
        \end{axis}
    \end{tikzpicture}
    \ref{globalLegend}
    \Description{Two plots comparing the coverage and error of two regression models. The left plot shows filled areas indicating error ranges, while the right plot shows line plots with a highlighted trade-off point.}
    \caption{The dilemma between coverage (proportion of predictions delivered) and error for two hypothetical regression models. Although Model 2 dominates Model 1 in most coverage levels, the goal is finding the best trade-off point between predictive error and coverage, minimizing the former while maximizing the latter, as shown in the right-side plot.}\label{fig:example}
\end{figure}
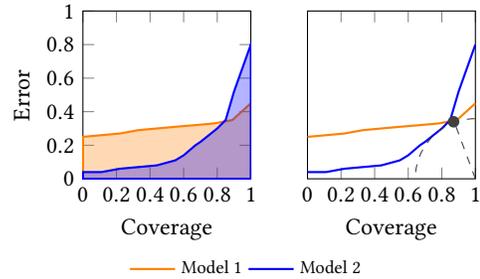

The growing use of artificial intelligence in society poses significant challenges to decision-making and the reliability of predictions due to the problem of uncertainty~\cite{dwivedi_artificial_2021}. For example, an AI system in healthcare could inaccurately predict patient outcomes, leading to incorrect treatment plans ~\cite{lu_improving_2022}. AI-based systems could inaccurately assess student performance in education, affecting their academic progress ~\cite{bailey2023ai}. Similarly, in finance, AI predictions could lead to incorrect credit risk assessments, impacting loan approvals \cite{redmond_data-driven_2002}. This demonstrates the cost of prediction errors and underscores the need for fairness, privacy, or reliability in predictive modeling across domains ~\cite{mehrabi_survey_2021}, and the difficulty in achieving it~\cite{Carvalho2023ThreeWayKnot}.

Selective regression (or ``reject option'') allows a model to abstain from a prediction in situations of high uncertainty ~\cite{chow_optimum_1957, chow_optimum_1970}, commonly measured by distribution-based proxies, such as conditional variance ~\cite{denis_active_2023, shah_selective_2022}. However, two critical gaps in the literature should be addressed. First, distribution-based proxies ignore inherent model biases, potentially leading to overconfidence in predictions ~\cite{gruber2023sources} and underestimating selective regression models' limitations ~\cite{lofstrom_bias_2015}. Second, the lack of a grounded evaluation approach results in a lack of consensus on assessing the trade-off between prediction performance and rejection rates (or coverage) ~\cite{chen_interpretable_2022, zaoui_regression_2020}, as illustrated in Figure~\ref{fig:example} concerning the most common evaluation metric in this context, the area under the curve (AUC). %However, it is irrelevant for a solution to dominate another for most coverage levels as long as the latter obtains the most optimal point for minimizing predictive error and maximizing coverage (as illustrated in Figure).

In this paper, we propose to leverage conformal prediction ~\cite{shafer_tutorial_2008} in selective regression tasks. Unlike distribution-based measures, conformal prediction builds calibrated prediction regions around the predicted value, given user-specified confidence levels, while guaranteeing that the true value falls within such regions with a certain probability ~\cite{shafer_tutorial_2008, vovk_algorithmic_2022, angelopoulos_uncertainty_2022}. We also propose a standardized approach to evaluate and compare selective regression approaches using a normalized distance-based framework %and its advantages w.r.t. using AUC.
in addition to AUC.

Our contributions can be summarized as follows:

\begin{enumerate}
    \item We introduce Conformalized Selective Regression \textbf{(CSR)}, a selective regression framework using conformal prediction, enhancing uncertainty measurement and model reliability.
    \item We propose an evaluation methodology to properly assess and compare selective regression methods based on their ability to achieve an optimal balance between predictive performance and model coverage. 
    \item Our results demonstrate that CSR consistently outperforms state-of-the-art methods, achieving a better balance between error rates and coverage across various domains. 
\end{enumerate}

\section{Related Work}

The ``reject option'' framework is fundamental for selective classification and regression by preventing incorrect predictions~\cite{chow_optimum_1957, chow_optimum_1970}. Selective classification has garnered significant attention~\cite{hellman_nearest_1970, de_stefano_reject_2000,kamath_selective_2020, el-yaniv_foundations_2010, geifman_selective_2017}, while selective regression introduces unique challenges~\cite{shah_selective_2022, geifman_selectivenet_2019, zaoui_regression_2020}. In practice, the ``reject option'' in selective regression allows models to abstain from predictions when uncertainty exceeds a predefined threshold. This mechanism is defined as:

\begin{equation}
\label{eq:selreg}
\Gamma_{\lambda}(X) =
\begin{cases}
f(X) & \text{if } u(X) \leq \lambda, \\
\textit{reject} & \text{otherwise}
\end{cases}
\end{equation}

\noindent where $\Gamma_{\lambda}(X)$ denotes the model's output, $f(X)$ the prediction for input $X$, $u(X)$ the uncertainty measure, and $\lambda$ the %\textcolor{blue}{user-defined} 
model's confidence level threshold necessary to make a prediction. 

Existing frameworks predominantly use conditional variance as a proxy for uncertainty. In regression analysis, conditional variance measures how much the predicted values can vary, represented as:

\begin{equation}
u(X) = \operatorname{Var}(Y \mid X) = \operatorname{E}\left[ (Y - \operatorname{E}(Y \mid X))^2 \mid X \right]
\end{equation}

%However, conditional variance may not account for the risks associated with heteroscedastic scenarios where error variance changes across inputs, potentially misleading users about the predictions' reliability.

However, conditional variance can be problematic in heteroscedastic scenarios where error variance changes across inputs, leading to inconsistent coverage. Our approach employs conformal prediction to provide adaptive confidence measures, accounting for varying uncertainty and potential model biases.  To address this, our approach employs conformal prediction to provide model-specific confidence measures, accounting for inherent biases in predictive models. Research supports conformal prediction's potential in diverse applications, from image classification to healthcare diagnostics \cite{angelopoulos_uncertainty_2022, lu_improving_2022}.

\subsection{Conformal prediction}

Conformal prediction, introduced by \cite{vovk_algorithmic_2022}, advances uncertainty quantification by providing statistically valid prediction intervals without assuming the underlying data distribution. %This methodology is applied across various domains to produce prediction intervals with a predefined confidence level. 
It is defined as:

\begin{equation}
C(X_{\text{test}}) = \{ y : s(X_{\text{test}}, y) \leq \hat{q} \}
\end{equation}

where \(C(X_{\text{test}})\) is the set of all possible outputs \(y\) such that the score function \(s(X_{\text{test}}, y)\) is less than or equal to a threshold \(\hat{q}\). The score function \(s(x, y)\) maps input-output pairs to real numbers \(\mathbb{R}\), with larger scores indicating worse agreement. Threshold $\hat{q}$ is the quantile of the calibration scores $s_1 = s(X_1, Y_1), \ldots, s_n = s(X_n, Y_n)$,

\[
\hat{q} = \frac{\lceil(n+1)(1-\alpha)\rceil}{n}
\]

\noindent ensuring the predictions are within a certain confidence level or accuracy defined by the conformal prediction framework. Here, $\alpha$ a the probability that the real value will be outside the conformal interval, i.e., lower $\alpha$, higher confidence, $n$ is the number of data points in the calibration set used to determine threshold $\hat{q}$.%, with a larger $n$ providing a more stable and reliable estimate. 

The goal is to achieve a coverage guarantee:
\begin{equation} 
\mathbb{P}\{Y_{n+1} \in C(X_{test})\} \geq 1-\alpha
\end{equation} 

where condition should hold for any joint distribution \( P_{XY} \) of the feature vectors \( X \), the response variables \( Y \), and any sample size \( n \). We do not estimate this probability directly; the conformal prediction framework guarantees coverage based on calibration and the chosen \( \alpha \), assuming exchangeability \cite{vovk_algorithmic_2022}.

\subsection{Conformalized Quantile Regression}

Quantile Regression, established by \cite{koenker_regression_1978}, is suitable for conformal prediction, which seeks not just point predictions but ranges that indicate where the true values might lie. It focuses on estimating conditional quantile functions and extends beyond mean predictions. Conformalized Quantile Regression (CQR) is developed in \cite{romano_conformalized_2019}. It builds prediction intervals with a predefined probability of encompassing the true response value, adapting to data heterogeneity. First, the data is split into a training $\mathcal{D}_{\text{train}}$ and calibration sets $\mathcal{D}_{\text{cal}}$. Then, two conditional quantile functions, \(q_{\hat{\alpha}_{\text{lo}}}\) and \(q_{\hat{\alpha}_{\text{hi}}}\), are fitted on the training set, capturing how lower and upper percentiles of the target variable change with different input features. Next, conformity scores on the calibration set quantify prediction interval errors. Given new input data \(X_{n+1}\), the prediction interval for \(Y_{n+1}\) is constructed by adjusting the estimated quantiles conformity-based scores, thus conformalizing the prediction interval.% This ensures prediction intervals are nearly perfectly calibrated.

%Selective regression was slightly mentioned within the conformal prediction framework \citep{angelopoulos_learn_2022}, marking an important step in exploring its implications. Though the study opened new avenues, it primarily introduced the concept and outlined potential directions for future research.

\section{Conformalized Selective Regression}
\label{sec:meth}
%\subsection{The Role of Conformalized  Interval Width in Selective Regression}
We introduce our Conformalized Selective Regression (CSR) framework, using conformal prediction to improve the reliability of selective regression by accounting for model-specific biases. %Conformal prediction creates prediction intervals that ensure the true value falls within them with high probability, enhancing uncertainty estimation. 
First, we find the conformity scores by calculating the intervals.

\begin{equation}
A_{cal} = max(y_{cal} - f_u(X_{cal}), f_l(X_{cal}) - y_{cal}),
\end{equation}

\begin{figure}[t!]
    \centering
    % First Plot
    \begin{tikzpicture}
        \begin{axis}[
            xlabel={Coverage},
            ylabel={Error},
            xmin=0, xmax=1,
            ymin=0, ymax=1,
            xtick={0, .2, .4, .6, .8, 1},
            ytick={0, 0.2, 0.4, 0.6, 0.8, 1},
            grid style=dashed,
            width=0.25\textwidth,
            height=0.25\textwidth,
            title={Model 1},
            tick label style={font=\scriptsize},
            label style={font=\small},
            title style={font=\small},
        ]
        % Model 1 line
        \addplot[color=red, thick] coordinates {
            (0.00, 1)
            (0.11, 0.7)
            (0.22, 0.5)
            (0.33, 0.4)
            (0.44, 0.3)
            (0.56, 0.22)
            (0.67, 0.19)
            (0.78, 0.18)
            (0.89, 0.19)
            (1.00, 0.2)
        };

        % Blue model
        \addplot[color=blue, thick] coordinates {
            (0.0, 0.04) (0.11, 0.04) (0.22, 0.06) 
            (0.33, 0.07) (0.44, 0.08) 
            (0.55, 0.11) (0.6, 0.14) 
            (0.67, 0.20) (0.7, 0.22) 
            (0.75, 0.26) (0.8, 0.30) 
            (0.85, 0.34) (0.9, 0.42) 
            (1.0, 0.80)
        };
        % Mark the closest point to (1,0)
        \node[circle,red, fill,inner sep=1pt] at (axis cs:0.99,0.2) {};
        % Euclidean distance illustration
        \draw[dashed, darkgray] (axis cs:1,0) -- (axis cs:0.99,0.2);
        \draw[dashed, darkgray] (axis cs:1,0) circle[radius={sqrt((1-0.99)^2 + (0-0.2)^2)}]; % Data units
        \end{axis}
    \end{tikzpicture}
    \hspace{0.5cm} % Adjust horizontal space between plots
    % Second Plot
    \begin{tikzpicture}
        \begin{axis}[
            xlabel={Coverage},
            ylabel=\empty,
            xmin=0, xmax=1,
            ymin=0, ymax=1,
            xtick={0, .2, .4, .6, .8, 1},
            ytick=\empty,
            grid=none,
            width=0.25\textwidth,
            height=0.25\textwidth,
            title={Model 2},
            tick label style={font=\scriptsize},
            label style={font=\small},
            title style={font=\small},
        ]
        % Model 2 line
        \addplot[color=green, thick] coordinates {
            (0.00, 0.02) 
            (0.15, 0.245)
            (0.20, 0.27)
            (0.26, 0.295)
            (0.33, 0.3)
            (0.38, 0.345)
            (0.44, 0.37)
            (0.50, 0.395)
            (0.56, 0.4)
            (0.61, 0.47)
            (0.67, 0.55)
            (0.72, 0.62)
            (0.78, 0.66)
            (0.80, 0.72)
            (0.83, 0.745)
            (0.89, 0.77)
            (0.94, 0.795)
            (0.98, 0.8)
            (1.00, 0.89)
         };

        % Blue model
        \addplot[color=blue, thick] coordinates {
            (0.0, 0.04) (0.11, 0.04) (0.22, 0.06) 
            (0.33, 0.07) (0.44, 0.08) 
            (0.55, 0.11) (0.6, 0.14) 
            (0.67, 0.20) (0.7, 0.22) 
            (0.75, 0.26) (0.8, 0.30) 
            (0.85, 0.34) (0.9, 0.42) 
            (1.0, 0.80)
        };
        % Mark the closest point to (1,0)
        \node[circle,blue, fill,inner sep=1pt] at (axis cs:0.75, 0.25) {};
        % Euclidean distance illustration
        \draw[dashed, darkgray] (axis cs:1,0) -- (axis cs:0.75,0.25);
        \draw[dashed, darkgray] (axis cs:1,0) circle[radius={sqrt((1-0.75)^2 + (0-0.25)^2)}]; % Data units
        \end{axis}
    \end{tikzpicture}
    \hspace{0.5cm} % Adjust horizontal space between plots
    % Third Plot
    \begin{tikzpicture}
        \begin{axis}[
            xlabel={Coverage},
            ylabel=\empty,
            xmin=0, xmax=1,
            ymin=0, ymax=1,
            xtick={0, .2, .4, .6, .8, 1},
            ytick=\empty,
            grid=none,
            width=0.25\textwidth,
            height=0.25\textwidth,
            title={Model 3},
            tick label style={font=\scriptsize},
            label style={font=\small},
            title style={font=\small},
        ]
        % Model 3 line
        \addplot[color=teal, thick] coordinates {
            (0.00, 0.01)
            (0.05, 0.02)
            (0.10, 0.02)
            (0.15, 0.02)
            (0.20, 0.02)
            (0.25, 0.02)
            (0.30, 0.02)
            (0.35, 0.02)
            (0.40, 0.02)
            (0.45, 0.02)
            (0.50, 0.02)
            (0.55, 0.02)
            (0.60, 0.02)
            (0.65, 0.04)
            (0.70, 0.1)
            (0.725, 0.15)
            (0.75, 0.2)
            (0.80, 0.34)
            (0.85, 0.5)
            (0.90, 0.7)
            (0.95, 0.8)
            (1.00, 0.9)
        };

        % Blue model
        \addplot[color=blue, thick] coordinates {
            (0.0, 0.04) (0.11, 0.04) (0.22, 0.06) 
            (0.33, 0.07) (0.44, 0.08) 
            (0.55, 0.11) (0.6, 0.14) 
            (0.67, 0.20) (0.7, 0.22) 
            (0.75, 0.26) (0.8, 0.30) 
            (0.85, 0.34) (0.9, 0.42) 
            (1.0, 0.80)
        };
        % Mark the closest point to (1,0)
        \node[circle,teal, fill,inner sep=1pt] at (axis cs:0.735, 0.15) {};
        % Euclidean distance illustration
        \draw[dashed, darkgray] (axis cs:1,0) -- (axis cs:0.735,0.15);
        \draw[dashed, darkgray] (axis cs:1,0) circle[radius={sqrt((1-0.735)^2 + (0-0.15)^2)}]; % Data units
        \end{axis}
    \end{tikzpicture}
    \Description{Three plots comparing the coverage and error of different regression models. Each plot shows two models, with the blue model present in all plots. The first plot (Model 1) uses red, the second plot (Model 2) uses green, and the third plot (Model 3) uses teal. Each plot highlights the point closest to the ideal zero-error, full-coverage point and illustrates the Euclidean distance from this point with dashed lines.}
    \caption{Comparison of simulated models based on coverage and error, including Euclidean distances from the ideal zero error point and full coverage. The plots illustrate that while a model might have a smaller AUC, it can still offer a more optimal trade-off between accuracy and coverage.}
    \label{fig:be}
\end{figure}
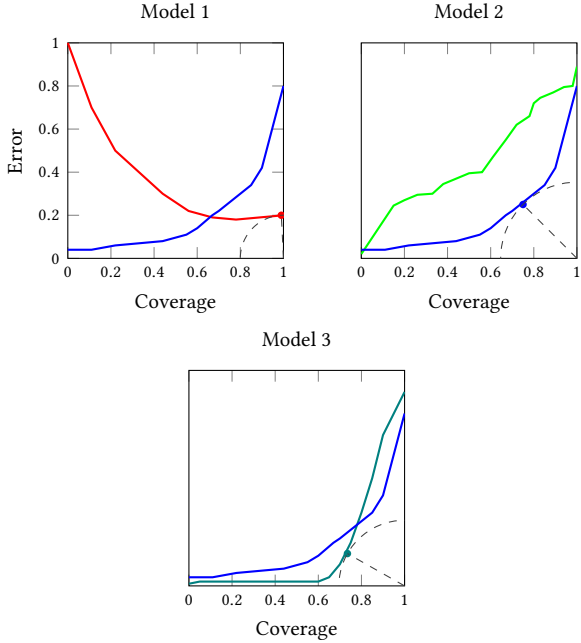

\noindent where \( y_{\text{cal}} \) is the calibration set of labels, \( X_{\text{cal}} \) the features in the calibration set, and \( f_l(X) \) and \( f_u(X) \) are the functions that predict the lower and upper ends of the non-conformalized prediction intervals. The calibration set ensures that any biases inherent in the \( f_l(X) \) and \( f_u(X) \) models are adjusted through the conformity scores, maintaining the reliability of the final prediction intervals. The training of these functions involves minimizing a quantile-based loss function, such as the pinball loss, on the training set \( \mathcal{D}_{\text{train}} \). Specifically, \( f_l(X) \) predicts the \( \alpha/2 \) quantile of dependent variable  that penalizes overestimates, while \( f_u(X) \) predicts the \( 1 - \alpha/2 \) quantile  that penalizes underestimates. This approach allows us to estimate the conditional quantiles of \( Y \) given \( X \) without needing explicit ground-truth bounds. For visualization of one point, we use the mean value of the interval. This process enables the models to accurately learn how the designated quantiles vary with the input features. Following the CQR procedure, we then compute:

\begin{algorithm}
\caption{Conformalized Selective Regression} 
\footnotesize
\label{alg:csr}
\begin{algorithmic}[1]

\REQUIRE Data set $\{(X_i, Y_i)\}_{i=1}^n$, with features $X_i \in \mathbb{R}^p$ and labels $Y_i \in \mathbb{R}$.
\REQUIRE Confidence level $\alpha \in (0, 1)$, rejection threshold $\lambda$.

\STATE Split data into training ($\mathcal{D}_{\mathrm{train}}$), calibration ($\mathcal{D}_{\mathrm{cal}}$), and test sets ($\mathcal{D}_{\mathrm{test}}$).
\STATE Train quantile regression models to predict the lower ($f_l(X)$) and upper bounds ($f_u(X)$) on $\mathcal{D}_{\mathrm{train}}$ for different quantile levels (e.g., \( \alpha/2 \) and \( 1-\alpha/2 \)).

\STATE Calculate scores: $A_{\mathrm{cal}} \gets \max(Y_i - f_u(X_i), f_l(X_i) - Y_i)$ for each $(X_i, Y_i)$ in $\mathcal{D}_{\mathrm{cal}}$.
\STATE Compute adaptive quantile threshold $\hat{q}_{\alpha}$: $\hat{q}_{\alpha} \gets \text{Quantile}\left(\frac{(n+1)(1 - \alpha)}{n}, A_{\mathrm{cal}}\right)$.

\STATE For each $(X_i, Y_i)$ in $\mathcal{D}_{\mathrm{test}}$, calculate the interval width: $W_{\alpha}(X_i) \gets f_u(X_i) - f_l(X_i) + 2\hat{q}_{\alpha}$; \COMMENT{We add \(2\hat{q}_{\alpha}\) to both bounds to symmetrically adjust the intervals.}

\STATE Predict or reject based on the width: If $W_{\alpha}(X_i) < \lambda$, predict $f(X_i)$; otherwise, reject (output 'No Prediction').

\RETURN Set of predictions or rejections for $\mathcal{D}_{\mathrm{test}}$.

\end{algorithmic}
\end{algorithm}

The conformalized interval width can be used as an uncertainty measure in selective regression models by setting $u(X) = W(x)$ as described in Eq.\ref{eq:selreg} -- wider interval indicates a higher level of uncertainty and vice-versa. As such, the model outputs a prediction \( f(X) \) only if the conformalized interval width is below threshold \( \lambda \). If not met, the model opts to reject the prediction. %\textcolor{blue}{which is automatically set based on the desired confidence level using calibration data, balancing coverage and prediction reliability}. %Beyond reflecting the potential range of outcomes, the conformalized interval width adapts dynamically based on the calibration set, offering a nuanced measure of the model's predictive uncertainty. 

%A . If we set an interval width threshold, the model can selectively make predictions only when the uncertainty is within the acceptable range. This allows the model to avoid making predictions when uncertainty is too high, thereby increasing the overall reliability of the model's output.
%Moreover, adapting the dynamic of the intervals via conformity scores from the calibration set ensures that the prediction intervals are not only statistically valid but also relevant in practice, catering to the peculiarities of the dataset at hand.

We apply Algorithm~\ref{alg:csr}, using a split conformal prediction framework, which partitions the data into distinct training and calibration subsets. This method involves training quantile regressors to establish preliminary bounds of the prediction interval, which are then refined using the calibration set to conform to the designated coverage criterion. However, the robustness of these predictive intervals must be rigorously tested to ensure their operational efficacy. This leads us to the critical aspect of evaluation in selective regression.

\begin{figure*}
\centering
% First line of figures: CSR vs Comparison Model 1 (Shah et al. 2022)
\includegraphics[width=0.85\textwidth]{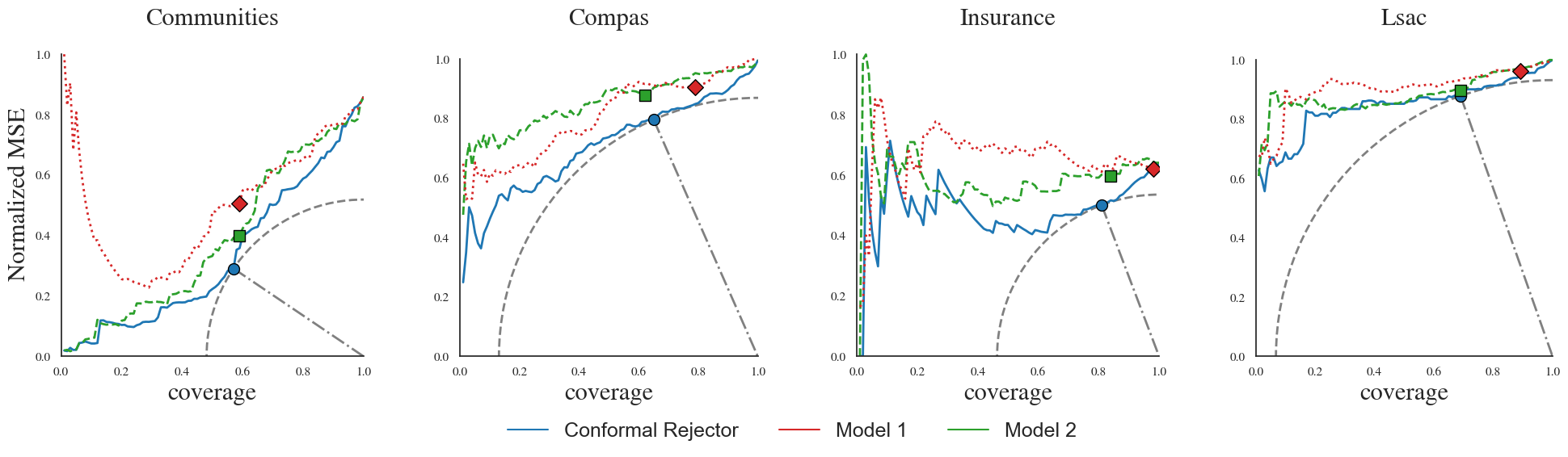}
\hfill

\caption{{Model Comparison on Normalized Mean Squared Error (nMSE) vs. Coverage with Random Forest Regressor.} This figure compares selective regression models across various datasets. Dashed and solid lines represent different modeling approaches, including CSR variants. Marks indicate optimal performance points - the best balance between prediction accuracy and coverage}
\label{1234}
\end{figure*}

\section{Evaluation in Selective Regression}

In selective regression, evaluating the predictive performance w.r.t. coverage is critical. Commonly used metrics, like AUC, do not necessarily shows the best effectiveness of such models (see Figure~\ref{fig:example}). %, as they fail to account for the rejected instances and hence do not contribute to the predictive performance. 
We propose to address not only this shortcoming in the current literature but also to propose an add-on standardized approach that can be used in calibration processes to anticipate ideal coverage levels using estimation methodologies, e.g. cross-validation \cite{Kohavi2001CV}. 

Our approach is as follows. For each model, we compute the normalized Mean Squared Error ($nMSE$) at each coverage level, using the maximum MSE across all models, and then find a point on this curve that is the closest to the ideal point of $\{nMSE=0, \text{Coverage}=1\}$. We then calculate the Euclidean distance of this point to the ideal point, i.e., a model that predicts all instances without error. By finding the model with the smallest Euclidean distance, we can identify which model provides the best trade-off between error rate and coverage.

%The goal of this experimental evaluation is two-fold. First, evaluate the performance of CSR models when compared to state-of-the-art selective regression approaches. Second, to demonstrate how existing evaluation approaches for selective regression are flawed while proposing a new standard approach to that aim.

The choice provides a straightforward measure of a model's effectiveness, becoming particularly valuable because it directly reflects the proximity of a model's performance to the optimal point, where the model would perfectly predict all instances without error while achieving full coverage. Models that minimize the Euclidean distance across varying levels of coverage are considered superior, as they are closer to achieving the ideal (and balanced) trade-off between predictive performance and coverage.

Figure~\ref{fig:be} illustrates these concepts by showing the position of each model relative to the ideal point on a plot, allowing for an intuitive understanding of which models offer the best trade-offs and thus should be prioritized in practical applications. The following analysis will dive deeper into these relationships, exploring how different models fare against each other in a comparative analysis using the Euclidean distance as a benchmark.

%It is important to mention that there are other metrics used for similar tasks, such as the Pareto Frontier. However, Euclidean distance provides a single value allowing straightforward ranking and comparison. It is easier to set thresholds or identify the nearest neighbor. In contrast, the Pareto Frontier is useful in multi-objective problems where improving one objective may worsen another. It highlights trade-offs rather than measuring direct distances, which can complicate the selection process. Each point on the Pareto Frontier is optimal based on different objective weights, making choosing a solution more complex.

While other metrics like the Pareto Frontier are used for similar tasks, Euclidean distance offers a straightforward single value, simplifying ranking and comparison. Unlike the Pareto Frontier, which is ideal for multi-objective problems and highlights trade-offs, Euclidean distance facilitates easier threshold setting and nearest neighbor identification, streamlining the selection process."

\section{Experiments and Results}

Our experimental evaluation aims to answer two research questions. First, is CSR more effective at finding optimal trade-offs between predictive performance and coverage w.r.t. other selective regression baselines? If so, second, are these results confirmed when restricting our search to higher coverage levels? In the following subsection, we present the state-of-the-art methods that serve as baselines, the data used, and other details. The source code is available here: \url{https://anonymous.4open.science/r/CSR_Submission-EDE6}

\subsection{Data}
In our experimental evaluation, we focused on four well-known datasets: COMPAS~\cite{barenstein_propublicas_2019}, Communities~\citep{redmond_data-driven_2002}, Insurance~\citep{lantz_machine_2019}, and LSAC~\cite{wightman1998lsac}. We split the data into training, calibration, and test sets (70\%, 10\%, 20\%). We used multiple regressors, such as Random Forest, XGboost, or Quantile Neural Networks, for a consistent comparison. This allowed a straightforward comparison between different rejection algorithms. 

\subsection{Baselines}

To benchmark the performance of CSR, we compared it against two state-of-the-art selective regression methods.

\subsubsection{Comparison Model 1: Fairness in Feature Representation~\cite{shah_selective_2022}}

This model ensures fairness through 'sufficiency in feature representation,' capturing all relevant information about sensitive attributes to promote fair predictions across groups. It calibrates mean and variance for consistency and uses a reject option when fairness criteria, specifically monotonic selective risk, are unmet. 

\subsubsection{Comparison Model 2: Plug-in $\epsilon$-Predictor with Reject Option~\cite{zaoui_regression_2020}}

The plug-in $\epsilon$-predictor with a reject option provides an optimal rule based on thresholding the conditional variance function and demonstrates a semi-supervised estimation procedure using the k-Nearest Neighbors (kNN) algorithm. It involves estimating the regression and variance functions and calibrating the rejection threshold using labeled and unlabeled datasets.

\subsection{Methods}

Using several models, including CSR with a fixed $\alpha = 0.05$, we calculate conformity scores from calibration subsets to estimate conformalized prediction intervals on the test set. Comparison Model 1 followed \citep{shah_selective_2022} specifications. For Comparison Model 2, we set $k = 10$ and repeated the procedure 100 times for stability. We evaluate the model’s performance across a range of  $\lambda$ values resulting in 0\% to 100\% coverage.

Comparing the performance of our proposed approach and state-of-the-art baselines is carried out on a model-by-model basis. As such, the set of predictions is fixed, i.e., all methods have the same performance at 100\% coverage. Then, we calculate the Euclidean distance to find the ideal point for each selective regression method, objectively assessing their ability to minimize errors and maximize coverage and guaranteeing that their performance is solely based on the effectiveness of their rejection strategies.

\subsection{Results}

Concerning the first question, comparing the CSR and other selective regression baselines w.r.t. their best predictive performance and coverage trade-off, Figure~\ref{1234} describes the performance of all three competing methods for the best Random Forest model in all datasets. Results show that CSR demonstrates the best trade-off between error rates and coverage, maintaining lower nMSE values. 

Based on these results, we look into our second research question, investigating if the previous results are confirmed when restricting our search to higher coverage levels. In practical applications, rejecting more than a certain percentage of predictions is often unrealistic. For example, achieving the best performance at 40\% coverage may be impractical. %As such, in Table~\ref{tab:rejector_performance}, we describe the performance of these methods for coverage values higher than 80\%, with the best models of each regressor used in this experimental evaluation. 
Results show that our proposed method consistently outperforms Model 1 and Model 2, demonstrating lower error rates across multiple datasets and coverage levels (0.8, 0.85, 0.9, and 0.95) in above 80\% cases. Further analysis of the model's performance, with coverage restricted to higher levels (0.8 -- 0.95), is provided in \url{https://anonymous.4open.science/r/CSR_Submission-EDE6} The results highlight CSR's effectiveness in providing reliable predictions while maintaining high coverage. To provide additional evidence, we analyzed an additional set of 25 regression datasets used in~\cite{ribeiro2020imbalanced} to validate our findings further. Here, results show that CSR achieves top performance in 80\% (20 datasets) of the cases, while Model 1 and Model 2 are the best options for 8\% (2 datasets) and 12\% (3 datasets), respectively.
\begin{table}[h]
\centering
\setlength{\tabcolsep}{4pt} %
\footnotesize % Reduce font size
\begin{tabular}{@{}lccc@{}}
\toprule
Dataset      & \multicolumn{1}{c}{Conformal} & Model 1 & Model 2 \\ 
              & \multicolumn{1}{c}{Rejector}  &         &         \\ \midrule
Communities  & \textbf{0.328}               & 0.505   & 0.381   \\
Compas       & \textbf{0.705}               & 0.800   & 0.846   \\
Insurance    & \textbf{0.484}               & 0.655   & 0.588   \\
Lsac         & \textbf{0.838}               & 0.897   & 0.877   \\ \bottomrule
\end{tabular}
\caption{Comparison of AUC scores}
\end{table}
Table1 presents the AUC values for various models across several datasets, where lower AUC values typically indicate better model performance in the context of this analysis. 

\section{Conclusion}

In this paper, we introduced CSR to enhance uncertainty measurement and model reliability in selective regression tasks. Our evaluation demonstrated that CSR outperforms existing methods by better balancing predictive accuracy and coverage across combinations of multiple data sets and models from distinct learning algorithms. In addition, we also proposed an evaluation approach that %overcomes the limitations of using AUC. 
 addresses the limitations of AUC, providing a more comprehensive assessment of model performance. Results highlight the potential of CSR in various domains, showing its effectiveness in managing uncertainty and addressing model-specific bias. Future work will explore further applications and refinements, such as using different underlying evaluation metrics and scenarios where predictive performance and coverage may have different weights.

%\citestyle{acmauthoryear}
\bibliographystyle{ACM-Reference-Format}
\bibliography{reference}

\end{document}